\documentclass[runningheads]{llncs}
\usepackage{geometry}

\usepackage{graphicx}
\usepackage{amsmath}
\usepackage{amssymb}
\usepackage{xspace}
\usepackage{xcolor}
\usepackage{dsfont}
\usepackage{rotating}
\usepackage{makecell}
\usepackage{multirow}
\usepackage{algpseudocode}
\usepackage{algorithm}
\usepackage{algorithmicx}
\usepackage[hidelinks]{hyperref}
\usepackage{paralist,url,doi,soul}
\usepackage{subcaption}

\usepackage[symbols,nogroupskip,sort=none]{glossaries-extra}

\glsxtrnewsymbol[description={dimension}]{d}{\ensuremath{d}}
\glsxtrnewsymbol[description={generation counter}]{g}{\ensuremath{g}}
\glsxtrnewsymbol[description={the number of parents}]{mu}{\ensuremath{\mu}}
\glsxtrnewsymbol[description={the number of offspring}]{lambda}{\ensuremath{\lambda}}
\glsxtrnewsymbol[description={objective function $\mathbb{R}^{d} \to \mathbb{R}$ (to be minimized)}]{f}{\ensuremath{f}}
\glsxtrnewsymbol[description={solution vector $\in \mathbb{R}^{d}$}]{x}{\ensuremath{\mathbf{x}}}
\glsxtrnewsymbol[description={i-th coordinate of solution vector $\mathbf{x}$}]{xi}{\ensuremath{\mathbf{x}_i}}
\glsxtrnewsymbol[description={i-th solution vector}]{ix}{\ensuremath{\mathbf{x}^{i}}}
\glsxtrnewsymbol[description={solution fitness}]{y}{\ensuremath{y}}
\glsxtrnewsymbol[description={Covariance matrix}]{C}{\ensuremath{\mathbf{C}}}
\glsxtrnewsymbol[description={center of mass}]{m}{\ensuremath{\mathbf{m}}}
\glsxtrnewsymbol[description={stepsize}]{sigma}{\ensuremath{\sigma}}
\glsxtrnewsymbol[description={Hill Valley Function}]{HV}{\ensuremath{\mathcal{HV}}}
\glsxtrnewsymbol[description={Redundancy function}]{R}{\ensuremath{\mathcal{R}}}

\begin{document}
\title{Avoiding Redundant Restarts in\\ Multimodal Global Optimization}
%
\author{Jacob de Nobel\inst{1},
Diederick Vermetten\inst{1}, 
Anna V. Kononova\inst{1}
Ofer M.~Shir\inst{2}, and Thomas B\"ack\inst{1}
}
\institute{
LIACS, Leiden University, Leiden, The Netherlands
\and 
Tel-Hai College and Migal Institute, Upper Galilee, Israel
}
\authorrunning{de Nobel, Vermetten, Kononova, Shir, and B\"ack}
%

%
\maketitle              
\begin{abstract}
Na{\"i}ve restarts of global optimization solvers when operating on multimodal search landscapes may resemble the Coupon's Collector Problem, with a potential to waste significant function evaluations budget on revisiting the same basins of attractions. In this paper, we assess the degree to which such ``duplicate restarts'' occur on standard multimodal benchmark functions, which defines the \textit{redundancy potential} of each particular landscape. We then propose a repelling mechanism to avoid such wasted restarts with the CMA-ES and investigate its efficacy on test cases with high redundancy potential compared to the standard restart mechanism.
\end{abstract}

\section{Introduction}

Finding an effective balance between exploring the domain and exploiting promising regions is one of the primary challenges when designing any iterative optimization heuristic, being subject to an underlying hard conflict. This design choice is especially important in multimodal search landscapes, where premature convergence leads to poor overall performance. 

Because of its challenging nature, a large number of algorithms have been developed for multimodal optimization~\cite{Preuss}, yet consisting of two main branches with different goals --- (i) niching methods~\cite{ShirNichingNATCOMP}, whose target is adjusted from locating the globally optimal solution to finding a wider set of high-quality optima, 
and (ii) ``upgraded global solvers'', which still target  
a single global optimum, but are better equipped to handle multimodality (see, e.g., \cite{hansencmamultimodal,auger2005restart}). 

While niching methods offer multiple approaches to treating such multimodal landscapes, primarily in promoting population diversity, these practices are usually not transferred to global optimization for various reasons (for instance, since focusing on solution diversity can hamper performance by slowing down convergence). Instead, global solvers often rely on rebalancing the search from global to local over time, e.g., by adjusting step sizes in evolutionary algorithms. 
To prevent convergence to sub-optimal solutions in these methods, restart mechanisms can be utilized to reset the algorithm's state, essentially starting a new optimization process by using the remaining function evaluation budget. This process is usually independent of the previous trajectory but can update the algorithm's strategy parameters, such as the population size within CMA-ES. 

However promising a restart scheme can be, it still carries the potential to encounter the equivalent of the \textbf{Coupon Collector's Problem}, which is rooted in the retrials' na{\"i}vity. Given $q$ unique coupons, the expected number of trials needed to collect them all, with replacement, is $q\cdot H_q$ (with $H_q:=\sum_{t=1}^{q} \frac{1}{t}$ being the $q^{th}$-harmonic number). 
Similarly, in the context of multimodal optimization, a straightforward approach of \emph{iteration} can be used to locate sequentially multiple peaks in the landscape by means of an \emph{iterative local search} (ILS)~\cite{lourencco2019iterated}.
If the procedure is blind to any information accumulated throughout previous runs, and it sequentially restarts stochastic search processes, the ambition to hit a different peak in every run resembles the collector's hope to obtain all the coupons in only $q$ trials. Overall, it is likely to encounter \emph{redundancy}, and the number of expected iterations is then increased by a factor. 
A {\bf redundancy factor} can be derived if the peaks are of equal height (an \textit{equi-fitness landscape}, i.e., the probability to converge into any of the $q$ peaks is uniform and equal to $1/q$), simply by normalizing the expected number of trials with respect to $q$ - that is an overall redundancy factor of $H_q$~\cite{ShirNichingNATCOMP}. Importantly, when dropping the equi-fitness assumption, this factor is expected to increase.

Mind should be given to this analogy and to the actual target of the ``collector'': the careful reader must note that a \emph{global} multimodal solver is not necessarily concerned with ``collecting'' all the optima (as in niching), since it is targeted at attaining only the best (i.e., picking only the ``top coupon''). At the same time, when operating in a black-box fashion, the global quality of such optima is often \textit{undecidable}. Therefore, the ``collector'' has no choice but to start a campaign of iterative restarts.

The idea of ``avoiding duplicates'' is a fundamental concept in heuristic search (e.g., the classical Tabu Search~\cite{Glover1998}), and even in statistical sampling (e.g., the conditioned Latin Hypercube Sampling that produces sampling designs while avoiding redundancy by accounting for external information~\cite{MinasnyMcBratney2006}). 
To investigate this idea in the context of multimodal optimization, we propose an alternative restart strategy in the context of the renowned CMA-ES algorithm. 

We provide the necessary background, especially concerning basins of attraction of local optima, and an illustrative motivation, in Section \ref{sec:problem}, and identify the potential evaluation redundancy problem caused by restarts in global optimization.
Then, we analyze the extent of this redundancy phenomenon on standard continuous landscapes, which motivates our proposal for an alternative restart strategy (Section \ref{sec:gain}). 
We propose and benchmark a restart strategy that integrates ideas from multimodal optimization, specifically \textit{repelling subpopulations in niching}, with the existing restart practices of CMA-ES. 
This new algorithm, the \emph{CMA-ES with repelling restarts} (RR-CMA-ES) is introduced in Section \ref{sec:RR-CMA-ES}.
This method is benchmarked on a wide set of problems, which shows that while wasted evaluations from ``duplicate'' restarts can be avoided, the precise methods to achieve this potential must be carefully calibrated to prevent deteriorating performance on some types of landscapes (Section \ref{sec:experiments}).
Conclusions and future work are discussed in Section \ref{sec:conclusions}.

\section{Preliminaries and Problem Formulation}\label{sec:problem}

\subsection{Basins of Attraction in Global Optimization}
We consider the \emph{global optimization} challenge of a single-objective, continuous 
minimization problem; the aim is to identify a \emph{single} solution $\vec{x}^* \in \mathcal{X} \subseteq \mathbb{R}^d$ from the feasible region $\mathcal{X}$, which minimizes a given objective function $f(\vec{x}): \mathcal{X} \rightarrow \mathbb{R}$:
\begin{equation}\label{eqn:problem}
\vec{x}^* = \arg \min_{\vec{x}\in \mathcal{X}} f(\vec{x})
\end{equation}
When the convexity property does not hold for $f$~\cite{Boyd}, it is often the case that other candidate solutions $\vec{x}_{\ell}\neq \vec{x}^{*}$ minimize certain neighborhoods of radii $\epsilon$, and thus each forms a \emph{local minimum}:
\begin{equation*}
\exists\epsilon>0\,~~~\forall\vec{x}\in \mathcal{X}\,\,:\,\,\left\Vert 
\vec{x}-\vec{x}_{\ell}\right\Vert <\epsilon\Rightarrow f(\vec{x}_{\ell})\leq f(\vec{x})
\end{equation*}
$\vec{x}^{*}$ is called the \textit{global optimizer}. Accordingly, the process of global optimization~\cite{globaloptimization} is considered successful when it concludes with locating it while escaping \textit{local optimizers} $\vec{x}_{\ell}$ (also referred to as \textit{traps}). 
Next, we would like to define the attraction basin of an optimizer (denoted as $\hat{\vec{x}}$, being either local or global), following~\cite{globaloptimization} (alternative definitions exist - see, e.g., \cite{AntonovSPIE2023}).
Given the standard Gradient Descent Algorithm~\cite{Boyd}, which is defined by its variation step from iteration $i$ to $i+1$ (with $\sigma(i)$ being the step-size), 
\begin{align}\label{eq:gradfbasin}
\displaystyle \vec{x}(i+1):= \vec{x}(i) -\sigma(i) \cdot \nabla f\left(\vec{x}(i)\right),    
\end{align}
we denote its initial search-point $\vec{x}(0):=\vec{x}_0$. 
The following set of points is defined whenever the limit $\lim_{i\rightarrow\infty}\vec{x}(i)$ exists:
\begin{align}
    \displaystyle \Omega = \left\{\vec{x} \in\mathbb{R}^{d}\left| \vec{x}(0) = \vec{x} 
\wedge \left.\vec{x}(i)\right|_{i\geq 0}~ \textrm{satisfies }\eqref{eq:gradfbasin} 
\wedge \lim_{i\rightarrow\infty}\vec{x}(i)~\textrm{exists} \right.\right\}
\end{align}
Then, given an optimizer $\hat{\vec{x}}$, we define its region of attraction using $\Omega$:
\begin{align}
\displaystyle \mathcal{A}(\hat{\vec{x}}) = \left\{\vec{x}\in\Omega \left| \vec{x}(0) = \vec{x} 
\wedge \left.\vec{x}(i)\right|_{i\geq 0}~ \textrm{satisfies }\eqref{eq:gradfbasin} 
\wedge
\lim_{i\rightarrow\infty}\vec{x}(i)=\hat{\vec{x}}\right.\right\}
\end{align}
Finally, and most importantly, the \emph{basin} of $\hat{\vec{x}}$ is the {\bf maximal level set} that is fully contained in $\mathcal{A}(\hat{\vec{x}})$.

We are particularly interested in addressing this global optimization challenge within black-box optimization, where the analytical form of the objective function $f$ is unknown to the solver and is thus treated as a black box, which, upon receiving input, yields a (continuous) output.

\subsection{CMA-ES}
The Covariance Matrix Adaptation Evolution Strategy (CMA-ES)~\cite{hansen2001completely} is a state-of-the-art method for single objective black-box optimization. The CMA-ES is a stochastic method that evolves a population of candidate solutions \gls{x} to an optimization problem $\gls{f}: \mathbb{R}^d \to \mathbb{R}$. To guide the search, it uses a parameterized multivariate normal distribution $\mathcal{N}(\gls{m}, \gls{sigma}\gls{C})$. The defining feature of the CMA-ES is that it can adapt the parameters of its mutation distribution to an arbitrary shape during optimization to guide the search. This makes it invariant to linear transformations of the search space and underpins its effectiveness on non-separable and ill-conditioned problems.

\subsection{Restart Mechanisms}

Restarts are commonly employed to prevent premature convergence in optimization algorithms. A whole class of algorithms relies on restart mechanisms to adapt local search methods to global optimization problems, dubbed as \textit{Iterative Local Search} (ILS)~\cite{lourencco2019iterated}. While ILS is mostly prevalent in discrete domains, techniques for continuous domains such as Multi-Level Single Linkage (MLSL)~\cite{MLSL} have shown promising performance on commonly used benchmark problems~\cite{pal2013benchmarking}.  
While restart methods are common in combination with local search, they can also be applied to more global optimizers, such as evolutionary algorithms. 
In its simplest form, for the CMA-ES, restarting involves resetting its mutation distribution to a standard multivariate normal distribution and setting the center of mass \gls{m} to a new location u.a.r. whenever \emph{local} restart criteria have been met. 
Additionally, the parameters of the algorithm can be altered to change its behavior following a restart, of which the most well-known strategies are IPOP~\cite{auger2005restart} and BIPOP~\cite{hansen2009benchmarking}. The IPOP-CMA-ES increases its population size by a factor of 2 on every consecutive restart. The BIPOP-CMA-ES alternates between two regimes; the first uses an increased population size and a larger value for the initial step size $\gls{sigma}^0$, while the second uses a smaller population size and $\gls{sigma}^0$. While other restart strategies have been proposed~\cite{loshchilov2012alternative,nishida2018benchmarking}, IPOP and BIPOP remain among the most commonly used methods. 

\subsection{Motivation}

While the aforementioned restart mechanisms have been shown to be effective on multimodal problems~\cite{hansen2009benchmarking}, no inherent method is implemented to ensure a subsequent restart does not converge to the same basin of attraction. By restarting u.a.r. it could very well be possible that several runs will be drawn to the same attractor. When this happens, the restart effectively spends more of the evaluation budget to potentially find an already identified solution. If this could be avoided by ensuring each consecutive restart only explores a previously unvisited region of the search space, we could potentially save a portion of the evaluation budget.

\begin{figure}[!tb]
    \centering
    \includegraphics[width=0.8\textwidth,trim={0 4mm 0 8mm},clip]{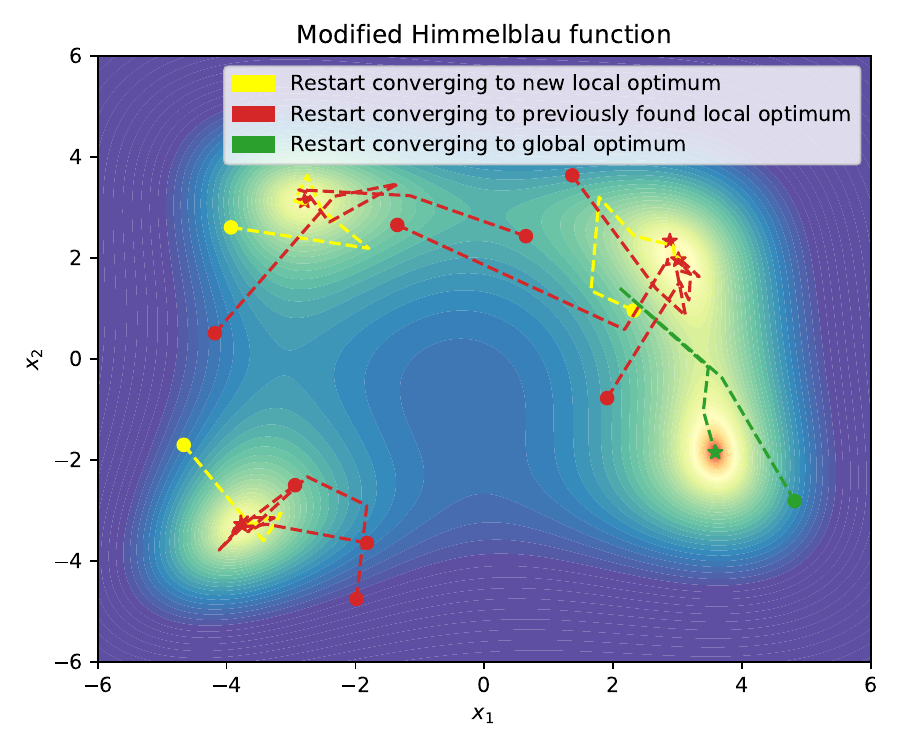}
    \caption{A single run of the CMA-ES with a simple u.a.r. restart strategy ($\gls{sigma}^0 = 2$), optimizing a modified version of the Himmelblau function ($f_{mh}$, see eq. \ref{eq:modhim}).     
    Every restart is visualized as a line, which shows the trace followed by the CMA-ES, where the circle shows the start of the trace and the star shows the finally obtained solution. The restart that converges to the global optimum is shown in green, and restarts converging to a new local optimum are shown in yellow. The traces in red show restarts that converge to local optima, which have already been found during previous restarts (yellow).}
    \label{fig:himmelblau}
\end{figure} 

Consider the example shown in Figure \ref{fig:himmelblau}, which shows a single run of the CMA-ES with a simple restart strategy solving a modified version of the 2-dimensional Himmelblau function: 
\begin{equation}
    f_{mh}(\gls{x}) = (x_1^2 + x_2 - 11) + (x_1 + x_2^2 - 7)^2 + \min\{0.01, ||\gls{x} - \gls{x}^*||_2 \}\label{eq:modhim}
\end{equation}
Here, $\gls{x}^*$ is the location of the global optimum, and $||\cdot||_2$ denotes the  Euclidean norm. From the figure, we can observe that the CMA-ES restarts several times during the optimization process.
Only during one of the restarts does the CMA-ES converge to the global optimum (green trace), while the other three local optima are found during other restarts. Notably, it can be observed that a considerable number of restarts are converging to local optima that have already been identified (red traces). In fact, this is the case for eight out of twelve restarts for this specific run. If we had been able to avoid revisiting previously found optima, considerable savings to the evaluation budget could have been realized. 
While one might argue that this is just a bad run or that a change to the initial $\sigma$ could avoid this type of behavior, we aim to demonstrate in the next section that this is not the case.

\subsection{Repelling Subpopulations}
The CMSA with \emph{Repelling Subpopulations} (RS-CMSA) \cite{rscmsa} has been proposed as a niching strategy that uses a number of parallel subpopulations to find multiple local optima in multimodal optimization. 
Our approach is inspired by the RS-CMSA, but it focuses on avoiding repeatedly sampling points in the basin of attraction of a local optimum that the algorithm has already identified.
The overall goal is to make the CMA-ES more sample-efficient for the goal of finding a single best solution (i.e., the best local optimum approximation, given a budget $B$ of function evaluations), rather than multiple solutions representing different local optima (i.e., multiple niches). 
The underlying methods, such as the usage of the Hill-Valley heuristic, restarts, and tabu regions, are inspired by the RS-CMSA.

\subsection{Hill-Valley Function as a Boolean Heuristic}
The Hill-Valley function $\texttt{HV}(\gls{x}^i, \gls{x}^j, \gls{f}) \in \{0, 1\}$, listed as Algorithm \ref{alg:hillvalley}, introduced as a part of the multinational EA~\cite{ursem1999multinational}, can be used as a heuristic to determine whether two points $\vec{x}^{i}$ and $\vec{x}^j$ belong to the same basin of attraction~\cite{maree2018real}. It calculates the objective function value for a maximum of $N_t = 10$ points on a line drawn between $\vec{x}^{i}$ and $\vec{x}^j$. If any of these points are of a higher objective function value, assuming minimization, a hill is assumed to be present between $\vec{x}^{i}$ and $\vec{x}^j$, which then do not share a basin of attraction. 

\begin{algorithm}
\caption{Hill-Valley test $\texttt{HV}(\gls{x}^i, \gls{x}^j, \gls{f})$}\label{alg:hillvalley}
\begin{algorithmic}
\Require $\vec{x}^{i} \in \mathbb{R}^{d}, \vec{x}^j \in \mathbb{R}^{d}, N_t \in \mathbb{N}, f: \mathbb{R}^{d} \to \mathbb{R}$
\For {$k \in 1, \dots, N_t$}
\State $\vec{x}^{test} = \vec{x}^{i} + \frac{k}{N_t + 1} (\vec{x}^j - \vec{x}^{i})$
\If {$\max\{{f(\vec{x}^{i}), f(\vec{x}^j)}\} \leq f(\vec{x}^{test})$}
\State  \Return 0
\EndIf
\EndFor
\State \Return 1
\end{algorithmic}
\end{algorithm}

\section{The Potential Gain of Avoiding Redundant Restarts}\label{sec:gain}

\subsection{Defining the Redundancy Measure}
Motivated by the \textit{redundancy factor} mentioned earlier in the context of the Coupon's Collector Problem, we would like to define a measure that quantifies the gain potential for having effective restarts. The trials of the Coupon's Collector must be adapted to the notion of restarts and to account for the global optimum, which constitutes the ``top priority coupon'' when following the analogy.
To this end, we define the \textit{restarts' redundancy factor} as the proportion of function evaluations spent by \emph{duplicate restarts} (different restarts that converged to a basin of attraction that was previously visited). 
This calculation excludes function evaluations spent within the global basin (regardless of being duplicates), denoted as $\gls{x}^*$ -- and it accumulates such function evaluations and normalizes them by the total budget.
With $r$ denoting the current number of restarts, having the current restart converging to a point $\gls{x}^{(r)}$, we use $\textrm{red}(\gls{x}^{(r)}) \in \{0, 1\}$ to determine whether the restart is redundant, relying on an oracle that answers the boolean query \texttt{sameBasin()} indicative of two points sharing a basin of the function $f$:
\begin{equation}\label{eq:redundant}
     \displaystyle \textrm{red}(\gls{x}^{(r)}) := \left[ \lnot \texttt{sameBasin}(\gls{x}^*, \gls{x}^{(r)}, \gls{f})) \right] \land \left[ \bigvee_{1 \leq k<r} 
     \hspace{-0.5em}
     \texttt{sameBasin}(\gls{x}^{(k)}, \gls{x}^{(r)}, \gls{f})\right] 
\end{equation}
Importantly, the previously defined Hill-Valley heuristic may approximate the boolean query \texttt{sameBasin()}, and will play this role in our implementation.

We then define the \textit{restarts' redundancy factor} as the normalized accumulated redundancy of a given set of restarts:
\begin{equation}\label{eq:potential}
    \displaystyle \textrm{RRF}\left(\left\{ \gls{x}^{(r)} \right\}_{r=1}^{R}  \right) := \frac{\sum_{r=1}^{R}\textrm{red}(\gls{x}^{(r)}) b^{(r)}}{B},
\end{equation}
where $b^{(r)}$ is the number of function evaluations spent by a restart, $R$ is the total number of completed restarts, and $B$ is the total budget of function evaluations spent.
Importantly, this factor accounts for a given set of restarts rather than measuring a search algorithm/mechanism (i.e., an implicit algorithmic measure).

\subsection{Numerical Assessment of Expected Redundancy Factors}\label{sec:redundancy}
We analyze the CMA-ES with three different restart strategies for several benchmark functions to identify the potential performance gains from avoiding duplicated restarts in global optimization. We compare a naive restart strategy (labeled `RESTART') alongside the popular IPOP and BIPOP restart strategies. Each strategy places the center of mass $\vec{m}$ uniformly at random in the domain on each restart and uses saturation for bound correction. The remaining settings of CMA-ES are left as default in the modCMA package~\cite{de2021tuning}. For each benchmark function, we perform 100 independent runs. In addition to logging the performance trajectories with IOHexperimenter~\cite{de2024iohexperimenter}, we log the center of mass and the number of evaluations used each time a restart is triggered. Based on this information, we can use Equation~\ref{eq:potential} to calculate the RRF for a given experiment.

\subsubsection{BBOB}
We benchmark on the noiseless, single-objective BBOB suite~\cite{bbobfunctions}. Originally proposed as part of the COCO benchmarking platform~\cite{hansen2020coco}, this set of 24 continuous optimization problems has been one of the last decade's most commonly used benchmarks for iterative optimization heuristics. For completeness, we perform our redundancy analysis on all 24 BBOB functions, even though a relatively large fraction consists of unimodal problems or problems with strong global structures where we don't expect to see any redundant restart. We run each problem for dimensionality $d\in\{2, 3, 4, 5, 6, 7, 8, 9, 10, 20\}$ for 10 instances. \\

 \begin{figure}[t]
    \centering
    \includegraphics[width=\textwidth,trim={0 6mm 0 9mm},clip]{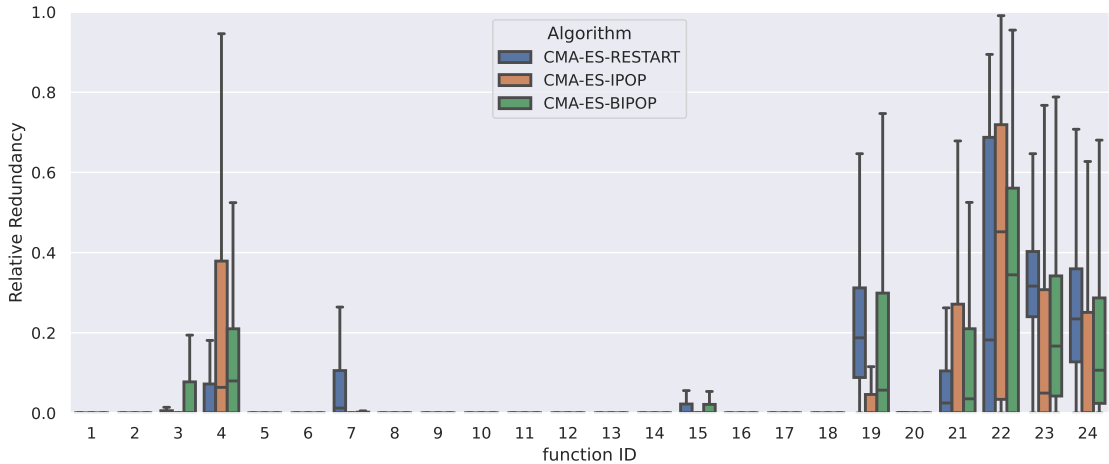}
    \caption{Boxplot showing the fraction of the total budget spent by restarts converging to a previously visited basin of attraction, the RRF (Equation~\ref{eq:potential}). All 24 objective functions from the BBOB benchmark are aggregated over all dimensions and instances for each tested restart strategy. }
    \label{fig:potential_bbob}
\end{figure}

Figure~\ref{fig:potential_bbob} shows the RRF for each function, aggregated over all dimensions, runs, and instances for each of the three restart strategies. From the figure, as expected, most redundant restarts occur in functions $f_{21}$--$f_{24}$, multimodal functions with weak global structure. We expect this is related to the weak structure since the chance we go to any given optima over one of the others should be related to the 'direction' of the function's global structure. Since there is no global structure and several local optima of equal height, we would be equally likely to end up in any of them randomly. \\

\begin{figure}[t]
    \centering
    \includegraphics[width=0.9\textwidth,trim={0 6mm 0 9mm},clip]{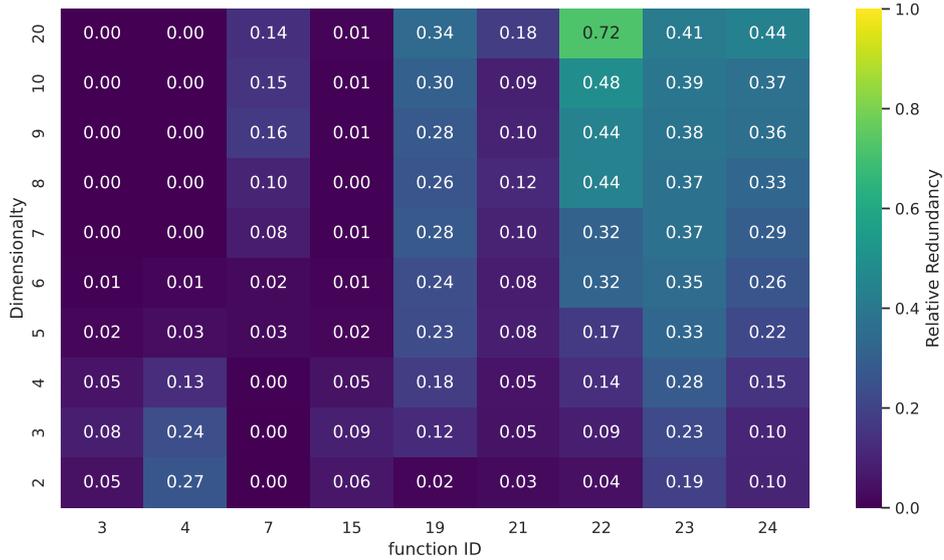}
    \caption{The average Relative Redundancy Factor over all instances for the CMA-ES using the `RESTART' strategy for the BBOB functions with any redundant restarts. The grid shows the RRF (Eq. ~\ref{eq:potential}) per dimension and function individually. }
    \label{fig:heatmap_restart_potential}
\end{figure}

The other functions showing some redundancy in Figure~\ref{fig:potential_bbob} are 
$f_3$, $f_4$, $f_{15}$ and $f_{19}$, which also exhibit some multimodality. Notably, $f_7$ is an interesting outlier, which, due to its plateauing landscape, causes several restarts for the `RESTART' strategy to be classified as redundant. Figure~\ref{fig:heatmap_restart_potential} provides a closer look at the functions with any redundant restarts, specifically showing the relation between dimensions for the `RESTART' strategy. Here we can observe that for some functions, such as $f_{19}$ and $f_{22}$, the RRF seems to increase with dimension; the opposite is true for functions  $f_{3}$ and $f_{4}$.

\subsubsection{CEC 2013}
\begin{figure}[!t]
    \centering
    \includegraphics[width=0.9\textwidth,trim={0 6mm 0 9mm},clip]{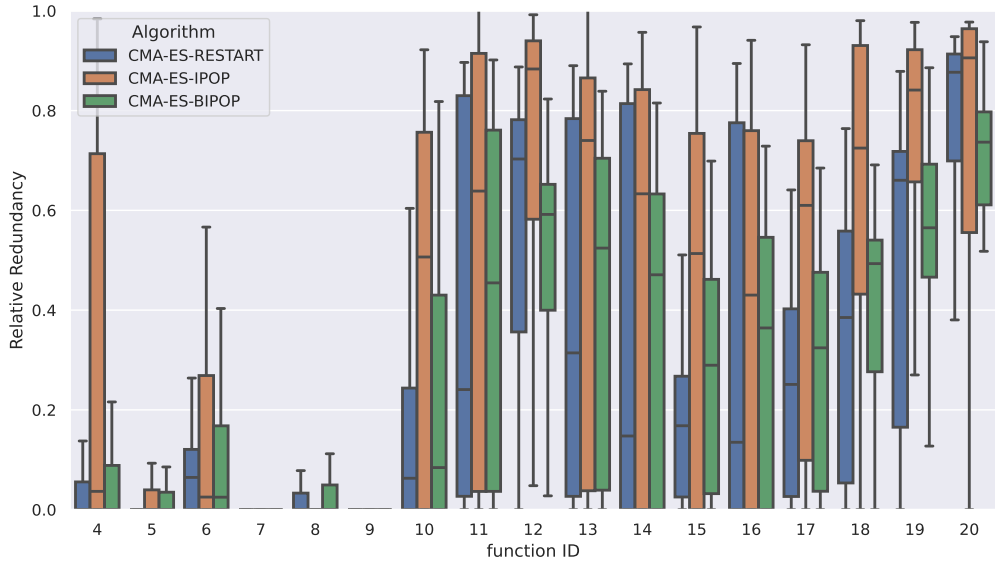}
    \caption{Fraction of budget which could be saved by avoiding convergence to redundant regions of the search space in subsequent restarts (as defined in Equation~\ref{eq:redundant}), for 16 functions from the CEC 2013 benchmark, aggregated over all instances for runs with the CMA-ES using different restart strategies. }
    \label{fig:potential_cec}
\end{figure}

While the overall potential to be gained from avoiding duplicated restarts is limited on the BBOB suite, we still see relatively large values in some multimodal problems. As such, we extend our setup to the CEC'13 suite~\cite{li2013benchmark}, which is designed specifically for multimodal optimization. We multiply each function with -1 to enable minimization. To transform the functions to global optimization problems, we select one of the existing global optima to turn into the only global one by adding a norm of distance as follows:
$f'(\vec{x}) = -f(\vec{x}) + \min(0.01, \|\vec{x}-\vec{x}^*\|^2)$. Given this ''globalization`` procedure, we create 10 instances of each of the 16 CEC problems (where each problem has a predefined dimensionality). The remainder of the setup is equivalent to the one used for the BBOB problems. The resulting redundancy factors are plotted in Figure~\ref{fig:potential_cec}, from which we observe that most problems defined for multimodal optimization show significant potential for saving redundant evaluations. We also observe that the distribution for IPOP is much wider and has a higher mean than the other methods. This happens because the population sizes are increased, leading to individual redundant runs containing a larger fraction of the total budget.  This also occurs to a lesser extent in BIPOP since the larger population sizes are interleaved with low population size restarts.

\section{Combating the Redundancy: Repelling CMA-ES}\label{sec:RR-CMA-ES}
Based on the RS-CMSA\cite{rscmsa}, which uses the concept of repelling subpopulations in a niching context, we introduce a CMA-ES with repelling restarts (RR-CMA-ES). Building on the standard restart strategies, the algorithm's idea is to define regions in the space where the CMA-ES cannot sample, as they were already visited during previous restarts. It maintains a set of tabu points $\vec{T}$, see Section~\ref{sec:tabu}, which describe a region. To determine the shape and size of this rejection region, we use the ideas from the RS-CMSA~\cite{rscmsa} and use the current covariance matrix $\mathbf{C}$ and step size $\sigma$. The size of the region is unique for each tabu point and controlled by $\delta(\vec{T})$. Algorithm \ref{alg:rrcma} gives a general overview of the method. 

\begin{algorithm}[!t]
\caption{RR-CMA-ES}\label{alg:rrcma}
\begin{algorithmic}[1]
\Require $\sigma_0 \in \mathbb{R}, \gamma \in \mathbb{R}, f: \mathbb{R}^{d} \to \mathbb{R}$
\State $\mathcal{T} \gets \emptyset$    \Comment{Tabu point archive}
\State $r \gets 0$
\While {not happy}
    \State $\sigma \gets \sigma_0, \; \mathbf{C} \gets \mathbf{I}, \; \vec{m} \sim \mathcal{U}(lb,ub)$

    \While {not any restart conditions}
        \State $\mathcal{X} \gets \emptyset, \; n_{\textrm{rej}} \gets 0$
        \While {$|\mathcal{X}| < \lambda$}
            \State $\vec{x} \gets \vec{m} + \mathcal{N}(0, \sigma\mathbf{C})$
            \If{$(\forall \vec{T} \in \mathcal{T}: \Delta_{\textrm{rej}}(\vec{x}, \vec{T}, \gamma, n_{\textrm{rej}}) = 0)$}    \Comment{Check if $\Vec{x}$ can be accepted}
                \State $\mathcal{X} \gets \mathcal{X} \cup \{\vec{x}\}$
            \Else
                \State  $n_{\textrm{rej}} \gets n_{\textrm{rej}} + 1$
                \Comment{Increase rejection count for $\Vec{m}$}
            \EndIf
        \EndWhile
        \State $\mathcal{F} \gets \texttt{evaluate}(\mathcal{X})$ \Comment{Evaluate all offspring}
        \State $\texttt{cmaUpdate} (\mathcal{X},\mathcal{F},\vec{m},  \sigma, \mathbf{C})$ \Comment{Continue the regular CMA-ES procedure}
    \EndWhile
    \State $r \gets r + 1$
    \State $\mathcal{T} \gets \mathcal{T} \cup \{\vec{m} \}$
    \Comment{Add $\Vec{m}$ to the tabu point archive}
\EndWhile
\end{algorithmic}
\end{algorithm}

\subsection{Tabu Points}
\label{sec:tabu}
The CMA-ES samples $\lambda$ points at every generation $g$. In the RR-CMA-ES, each newly sampled point $\vec{x}$ is tested against the archive $\mathcal{T}$ of tabu points before being accepted for evaluation (line 9). 
A tabu point defines a location in the search space where the optimizer is not allowed to go. While initially imagined for combinatorial optimization, for the continuous spaces the CMA-ES deals with, it defines a hyper-ellipsoid centered around a point where the optimizer cannot sample. Specifically, a tabu point $\vec{T}$ consists of a triplet $(\vec{x}^{\vec{T}}, f(\vec{x}^{\vec{T}}), n^{\vec{T}})$, where $\vec{x}^{\vec{T}}$ is the location in the search space with $f(\vec{x}^{\vec{T}})$ its corresponding fitness and $n^{\vec{T}}$ being the number of times the CMA-ES has converged to $\vec{T}$ during previous restarts. 
During sampling, a tabu point $\vec{T}$ rejects a newly sampled point $\vec{x}$ according to the following boolean query ($\Delta_{\textrm{rej}} \in \{0,1\} $): 
\begin{equation}\label{eq:reject}
   \Delta_{\textrm{rej}}(\vec{x}, \vec{T}, \gamma, n_{\textrm{rej}}) =  \left(\frac{d_m(\vec{x}, \vec{x}^{\vec{T}}, \mathbf{C}^{-1})}{\sigma}  < \gamma^{n_{\textrm{rej}}} \delta(\vec{T}) \right)
\end{equation}
Here, $d_m$ denotes the Mahanolobis distance metric, scaled by the current covariance matrix $\mathbf{C}$ and step size $\sigma$, and $\delta(\Vec{T})$ denotes the rejection radius around the tabu point $\Vec{T}$. To avoid stagnation, a shrinkage factor $0 < \gamma < 1$ is applied to $\delta(\vec{T})$ with $n_{\textrm{rej}}$ denoting the number of times a point has been rejected in the current generation. Then for every newly sampled point $\vec{x}$ it is accepted if and only if: 
\begin{equation}
    \forall \vec{T} \in \mathcal{T}: \Delta_{\textrm{rej}}(\vec{x}, \vec{T}, \gamma, n_{\textrm{rej}}) = 0 
\end{equation}

\subsection{Restarting}
Upon every restart $r$, the archive $\mathcal{T}$ is updated with the converged center of mass $\vec{m}$. Using the Hill Valley $\texttt{HV}$ routine, we check whether $\vec{m}$ has converged to a new basin of attraction. If the restart converged to a new basin of attraction, that is $\texttt{HV}(\vec{m}, \vec{x}^{\vec{T}}, f) = 0, \forall \vec{T} \in \mathcal{T}$, the new tabu point ($\vec{m}, f(\vec{m}), 1)$ gets added to the archive. If the restart converged to a point that is already present in $\mathcal{T}$, $n^{\vec{T}}$ gets increased by 1, and if $f(\vec{m}) < f(\vec{x}^{\vec{T}})$, $(\vec{m}, f(\vec{m}), n^{\vec{T}})$ replaces $(\vec{x}^{\vec{T}}, f(\vec{x}^{\vec{T}}), n^{\vec{T}})$. 

\subsection{Search space coverage}
We define a \textit{coverage factor} $c$, which controls the ratio of the total volume of the search space $S = \prod_{i=1}^d (ub_i - lb_i)$ that is covered by the repelling regions of all tabu points. 
Intuitively, $\frac{1}{c}$ denotes the maximal proportion of $S$, which is unavailable for the CMA-ES to sample. 
This is divided amongst all tabu points, where points with a higher $n^{\vec{T}}$ cover a larger part of this volume. Then for each tabu point, the volume of the repelling region, normalized for $\sigma_0$ is:

\begin{equation}
    V(\vec{T}) = n^{\vec{T}} \frac{S}{c \sigma_0 R}
\end{equation}
where $R$ denotes the total number of restarts. This is used to calculate the rejection radius: 
\begin{equation}
    \delta(\vec{T}) = V(\vec{T})^{\frac{1}{d}} \frac{\Gamma(\frac{d}{2} + 1)^\frac{1}{d} }{\sqrt{\pi}}
\end{equation}
where $d$ denotes the dimensionality and $\Gamma(\cdot)$ the gamma function.

\section{Proof-of-Concept: Repelling CMA-ES in Action}\label{sec:experiments}

\begin{figure}[t]
\centering
\includegraphics[width=\textwidth,trim={0 6mm 0 9mm},clip]{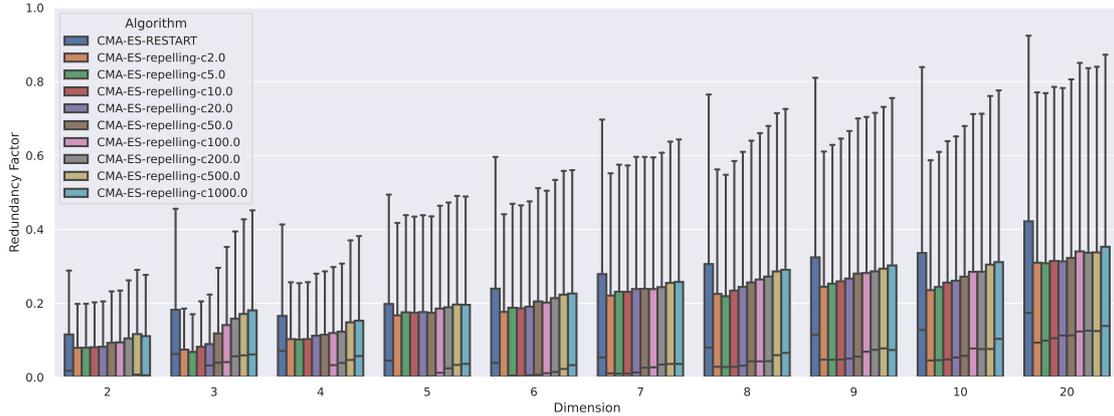}
\caption{Distribution of redundant function evaluation over the BBOB functions where redundant restarts were found in Section~\ref{sec:redundancy}, i.e. $f_3, f_3, f_{15}, f_{19}, f_{21}, f_{22}, f_{23}$ and $f_{24}$, separated by problem dimensionality. The CMA-ES with the `RESTART' strategy is compared to the RR-CMA-ES, with different coverage factors $c$. }
\label{fig:bbob_potential_saved}
\end{figure}

To illustrate the workings of the RR-CMA-ES, we benchmark several versions with redundancy factors ranging from 2 to 1000 on both the BBOB and CEC'13 functions. We perform $50$ independent runs on each of the $10$ instances used in Section~\ref{sec:redundancy} and compare the results to the original CMA-ES. In this section, we use the versions with the default restart mechanism, but the IPOP and BIPOP results are available in our reproducibility repository~\cite{reproducibility_and_figures}. To gauge whether the repelling strategy prevents duplicate restarts from occurring, we perform the same expected redundancy factor calculation from Equation~\ref{eq:potential} on the runs from the RR-CMA-ES, and visualize the results in Figure~\ref{fig:bbob_potential_saved}. From this figure, we can see that on average, the RR-CMA-ES with the lowest coverage factor (the largest repelling regions) quite effectively prevents different restarts from converging to the same basins. \\
\begin{figure}[t]
\centering
\includegraphics[width=\textwidth]{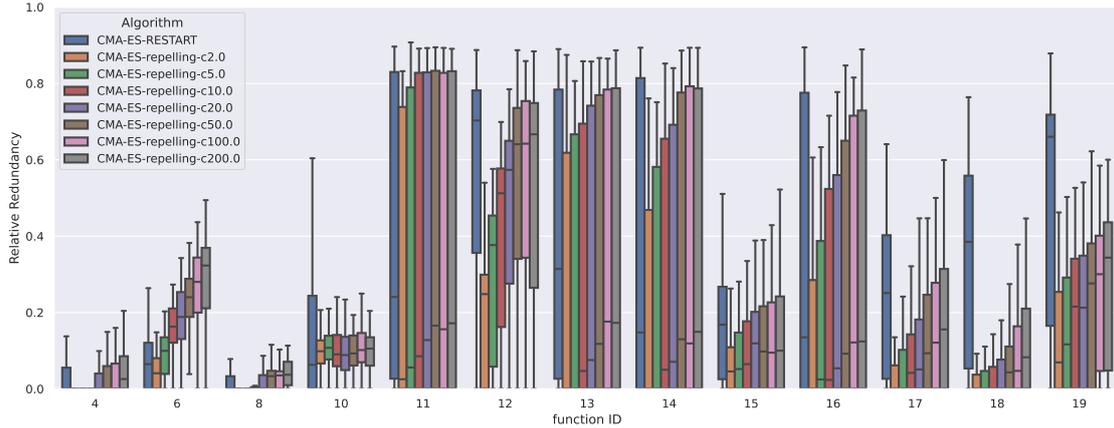}
\caption{Distribution of redundant function evaluation over the CEC'13 functions, only function with any redundant restarts for CMA-ES-RESTART are shown. The CMA-ES with the `RESTART' strategy is compared to the RR-CMA-ES, with different coverage factors $c$.}
\label{fig:cec_potential_saved}
\end{figure}

We can perform the same redundancy-based comparisons for the CEC functions, where in  Figure~\ref{fig:cec_potential_saved} we see how the potential changes when the repelling strategy is used. From this figure, we can see that the impact of the repelling regions is even larger than that of the BBOB problems, but with large variations between the different functions. However, the relation to the coverage factor seems consistent, with a lower factor leading to fewer redundant restarts. 

\begin{figure}[t]
    \centering
    \begin{subfigure}[][][t]{0.32\textwidth}
    \includegraphics[width=\textwidth]{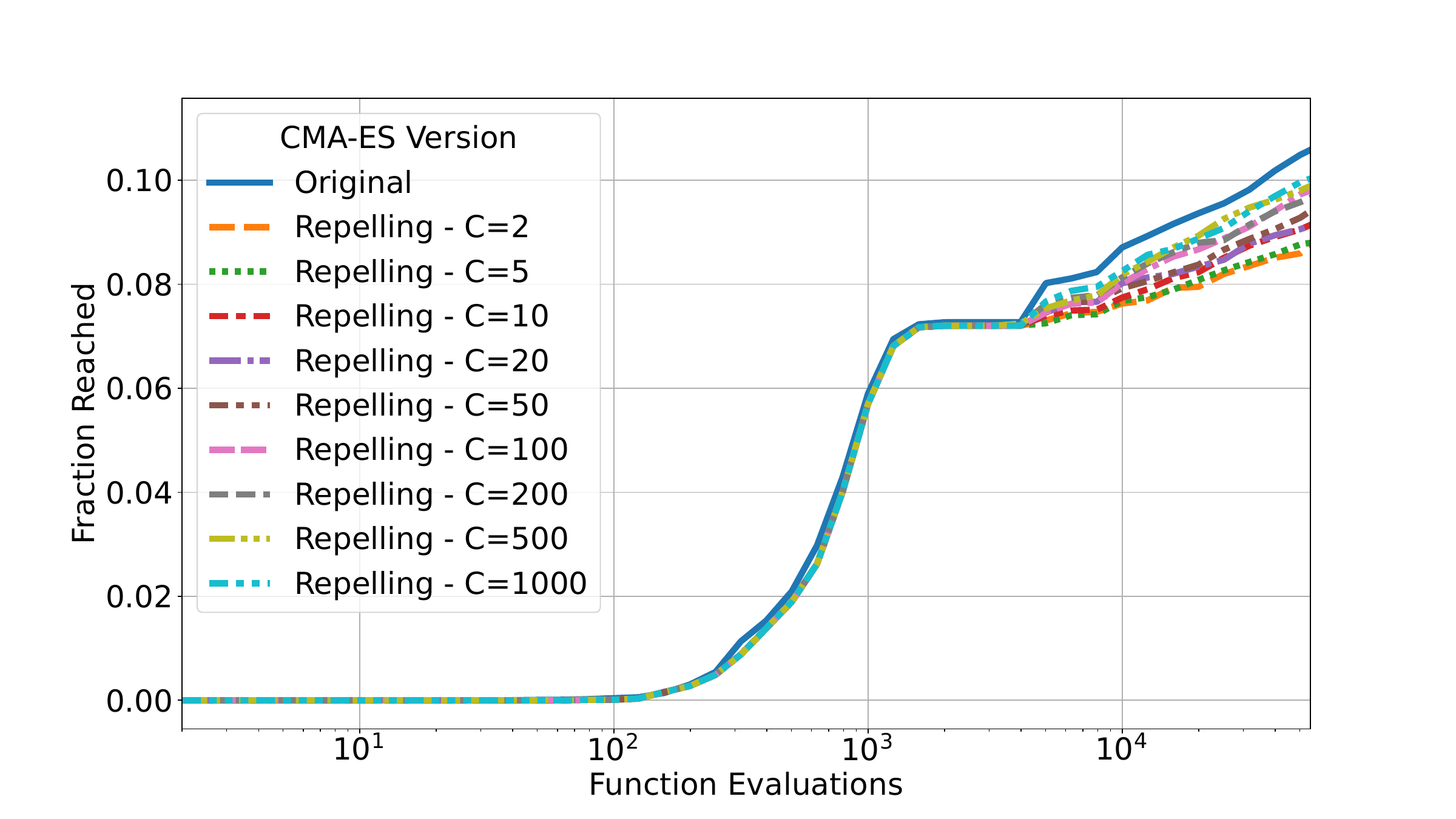}
    \caption{F4}
    \label{fig:ecdf_f4_10d}
    \end{subfigure}
    \begin{subfigure}[][][t]{0.32\textwidth}
    \includegraphics[width=\textwidth]{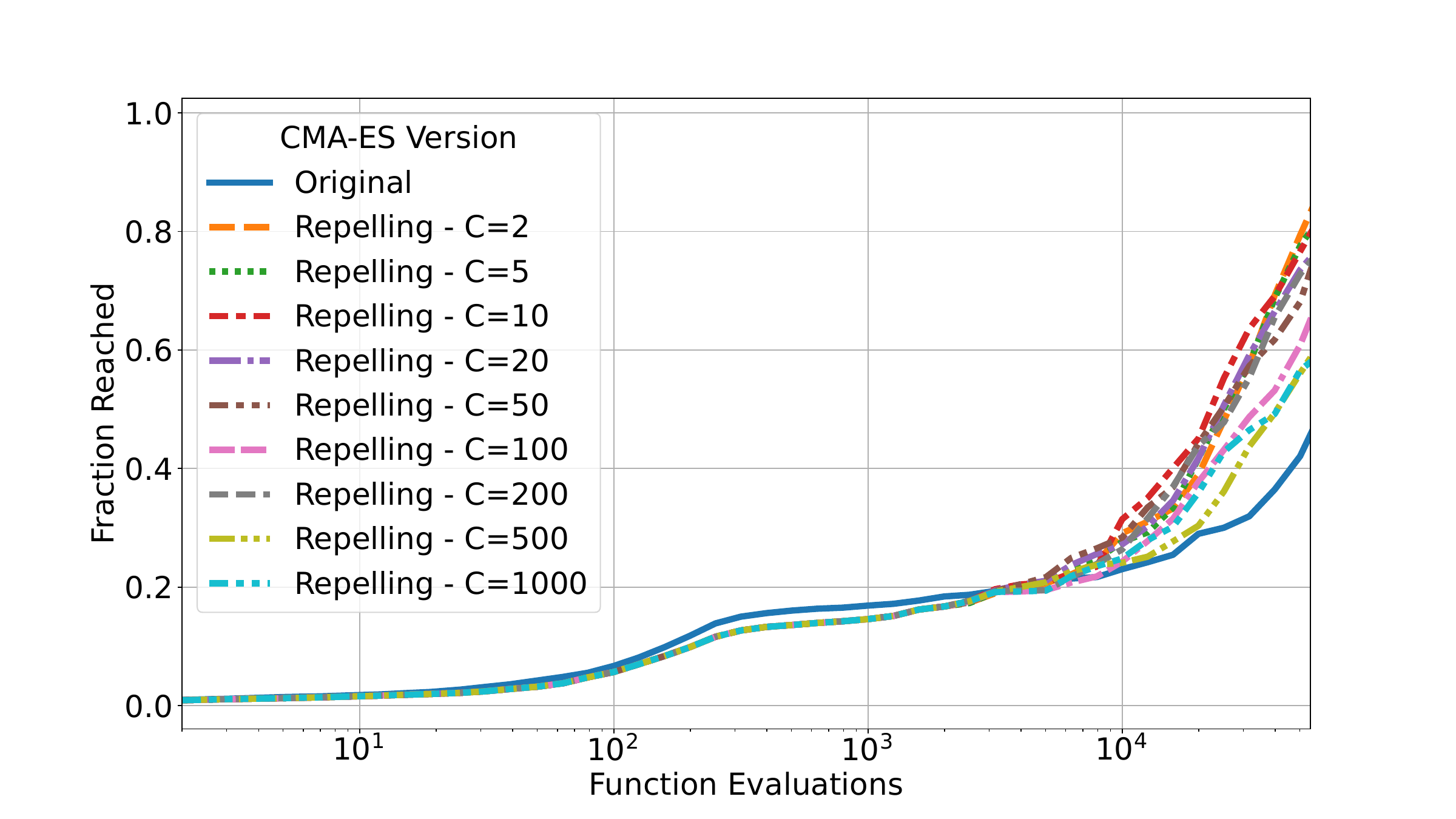}
    \caption{F22}
    \label{fig:ecdf_f22_10d}
    \end{subfigure}
    \begin{subfigure}[][][t]{0.32\textwidth}
    \includegraphics[width=\textwidth]{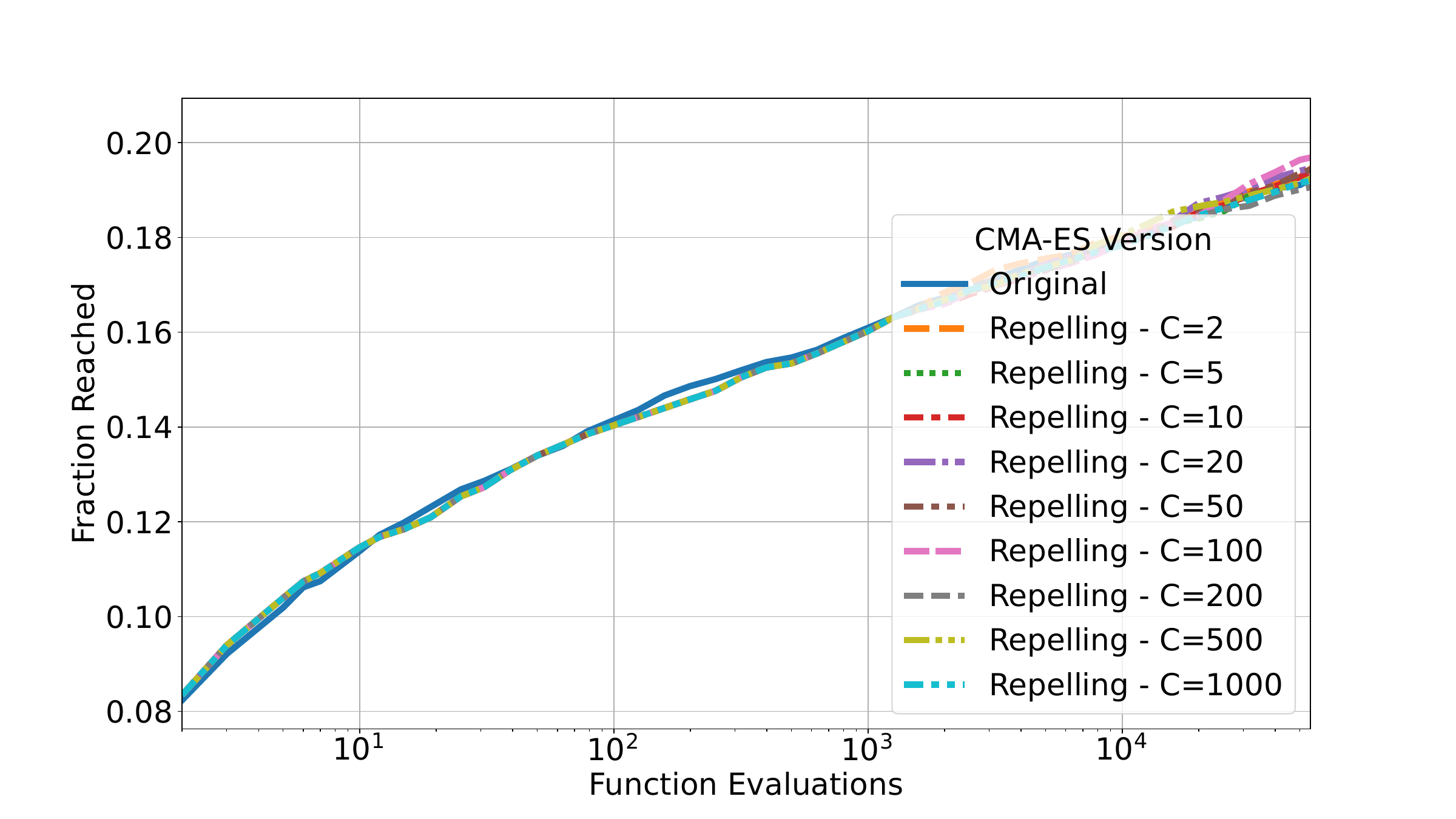}
    \caption{F23}
    \label{fig:ecdf_f23_10d}
    \end{subfigure}
    \caption{ECDF of the different versions of RR-CMA-ES compared to the original CMA-ES, for selected BBOB functions (in dimensionality 10) where potential improvement was observed from Figure~\ref{fig:potential_bbob}.}\label{fig:ecdf}
\end{figure}

Given the promising reduction in redundant evaluations, we look at the performance of the considered algorithm variants. In Figure~\ref{fig:ecdf}, we show the empirical cumulative distribution plot\footnote{This ECDF is based on the empirical attainment function, equivalent to infinite targets for the standard ECDF~\cite{lopez2024using}.} on several BBOB problems where potential gains were identified. We use bounds $10^2$ and $10^{-8}$ with log-scaling between them to be consistent with the common COCO setup~\cite{hansen2020coco}. This figure shows no difference between these methods in the early stage of the search (since no restarts have been triggered yet), but the repelling can increase or decrease overall performance depending on the function. 
An important aspect to note is that on functions with a clear global structure, such as F4, the repelling regions might adversely affect convergence, as shown in Figure~\ref{fig:ecdf_f4_10d}. Thus, a balance between preventing redundant restarts and avoiding steering the search away from regions near the global optimum has to be found.

\section{Conclusions and Future Work}\label{sec:conclusions}
In this paper, we reflected that global optimization of multimodal problems is an area where current optimization algorithms can still be improved in light of an evident redundancy of restart mechanisms. When independent restarts converge to the same local optima, these essentially waste function evaluations, which could be prevented by more effectively guiding the search in subsequent restarts.  

Our proposed repelling-based restart method for CMA-ES shows a slight performance improvement over the default restart mechanisms for specific cases and reduces redundant convergence for the BBOB functions. For the multimodal CEC functions specifically, the method shows a large reduction of redundant restarts compared to a standard restart strategy. 

Further research into the precise ways of avoiding these redundancies is still required. In particular, the rejection criteria of the tabu regions could be modified to incorporate information about the fitness values found in this region to better handle global structure. 

The tabu regions' shapes could also be formed based on the converged covariance matrix at the point or slightly before the restart is triggered to potentially better capture the precise basin shape. Furthermore, extended practices such as \textit{regularization} could also be exercised to remedy numerical issues and to obtain a more accurate structure (see, e.g., \cite{Shir-FOCAL}), relying on the theoretical relation of the ESs' covariance matrix and the landscape local Hessian \cite{Shir-Yehudayoff_TCS2020}.   

In addition to more effectively preventing a restart from converging to a known basin of attraction, one could also utilize the information from prior restarts to more intelligently select the new location and initial parameterization of new restarts. In comparison to, e.g., MLSL, starting points could be selected based on which regions of the domain have already been sampled, which would also cause the repelling mechanism to trigger less often during the initial phase of the restarted run. 

In the context of global optimization, the explicit criteria used to trigger a restart mechanism also play a large role in the algorithm's performance. Depending on the optimization goal, it might be worthwhile to restart more frequently and to exploit the best-attained basins of attraction using a local search method with a small fraction of the total budget.

\section*{Acknowledgements}

%
%
%
\bibliographystyle{splncs04}
\bibliography{references}
%

\end{document}